\documentclass[sigconf]{acmart} % Camera-ready version

\AtBeginDocument{%
  \providecommand\BibTeX{{%
    \normalfont B\kern-0.5em{\scshape i\kern-0.25em b}\kern-0.8em\TeX}}}

\copyrightyear{2022}
\acmYear{2022}
\setcopyright{rightsretained}

\acmConference[KDD '22]{Proceedings of the 28th ACM SIGKDD Conference on Knowledge Discovery and Data Mining}{August 14--18, 2022}{Washington, DC, USA}
\acmBooktitle{Proceedings of the 28th ACM SIGKDD Conference on Knowledge Discovery and Data Mining (KDD '22), August 14--18, 2022, Washington, DC, USA}
\acmISBN{978-1-4503-9385-0/22/08}
\acmDOI{10.1145/3534678.3542681}
\usepackage{etoolbox}
\makeatletter
\patchcmd{\maketitle}{\@copyrightpermission}{
   \begin{minipage}{0.3\columnwidth}
     \href{https://creativecommons.org/licenses/by/4.0/}{\includegraphics[width=0.90\textwidth]{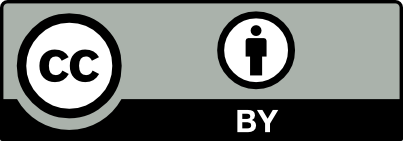}}
   \end{minipage}\hfill
   \begin{minipage}{0.7\columnwidth}
     \href{https://creativecommons.org/licenses/by/4.0/}{This work is licensed under a Creative Commons Attribution International 4.0 License.}
   \end{minipage}
 
   \vspace{5pt}
}{}{}

\makeatother
\acmSubmissionID{health21}

\newcommand\theapp{Safe Delivery App}
\newcommand\theorg{Maternity Foundation}
\newcommand\citetheapp{\cite{sda}}
\newcommand\citetheorg{\cite{mf}}
\newcommand\theotherorg{, the University of Copenhagen, and the University of Southern Denmark}

\begin{document}

%\title{Personalization and User Engagement with Mobile Health Applications}
%\title{Towards a Full Assessment of User Engagement and Churn with Mobile Health Applications}
\title{User Engagement in Mobile Health Applications}


%\affiliation{%
%  \institution{benshi.ai}
%  \streetaddress{Passeig de Gràcia, 74}
%  \city{Barcelona}
%\state{Catalunya}
%  \country{Spain}
%  \postcode{08008}
%}

%\email{ana@benshi.ai}
%\affiliation{%
%  \institution{benshi.ai}
%  \streetaddress{Passeig de Gràcia, 74}
%  \city{Barcelona}
%\state{Catalunya}
%  \country{Spain}
%  \postcode{08008}
%}
%\orcid{0000-0003-1465-7065}

\author{Babaniyi Yusuf Olaniyi}
\email{babaniyi,ana,africa@benshi.ai}
\author{Ana Fernández del Río}
\author{África Periáñez}
\affiliation{%
  \institution{benshi.ai}
  \streetaddress{Passeig de Gràcia, 74}
  \city{Barcelona}
  \country{Spain}
  \postcode{08008}
}

\author{Lauren Bellhouse}
\email{lauren@maternity.dk}
\affiliation{%
  \institution{Maternity Foundation}
  \streetaddress{Forbindelsesvej 3}
  \city{Copenhagen}
  \country{Denmark}}

\renewcommand{\shortauthors}{Babaniyi Yusuf Olaniyi et al.}

%%\renewcommand{\shortauthors}{Trovato and Tobin, et al.}

\begin{abstract}
Mobile health apps are revolutionizing the healthcare ecosystem by improving communication, efficiency, and quality of service. In low- and middle-income countries, they also play a unique role as a source of information about health outcomes and behaviors of patients and healthcare workers, while providing a suitable channel to deliver both personalized and collective policy interventions. We propose a framework to study user engagement with mobile health, focusing on healthcare workers and digital health apps designed to support them in resource-poor settings. The behavioral logs produced by these apps can be transformed into daily time series characterizing each user's activity. We use probabilistic and survival analysis to build multiple personalized measures of meaningful engagement, which could serve to tailor contents and digital interventions suiting each health worker's specific needs. Special attention is given to the problem of detecting churn, understood as a marker of complete disengagement. We discuss the application of our methods to the Indian and Ethiopian users of the \theapp{}, a capacity building tool for skilled birth attendants. This work represents an important step towards a full characterization of user engagement in mobile health applications, which can significantly enhance the abilities of health workers and, ultimately, save lives.%\looseness=-1
\end{abstract}




\maketitle

% TODO: FINAL REVIEW ON REFERENCES THROUGHOUT MAIN TEXT TO SUPPLEMENTARY MATERIAL

\section{Introduction}
\label{sec:intro}

Digital health solutions have disruptive potential for precision public health and improved health outcomes \cite{dowell2016, dolley2018, buckeridge2020, Marsch2021, Overdijkink2018}, as they allow for the collection of large quantities of behavioral and health-related data that can be used to design and evaluate evidence-based policies. 
%to improve health outcomes. 
%In resource poor settings in particular \cite{Hosny2019, Wahl2018}, driven by the growing mobile penetration in low and middle income countries (LMICs), there is an 
In particular, in low- and middle-income countries~\cite{Hosny2019, Wahl2018}, the growing mobile penetration has led to the deployment of an increasing number of digital tools to assist patients and healthcare providers
%\cite{Abraha2017, Gimbel2018, Feroz2017} 
\cite{Gimbel2018, Feroz2017} 
and help them overcome the scarcity of resources. Such mobile health solutions include capacity building tools, apps that provide healthcare workers with needed supplies (e.g. drug delivery services) or medical resources (e.g. arrangement of test appointments), tools that connect them to patients (e.g. for a remote follow-up) or physicians (e.g. to get their opinion on test results), and even apps to assist them in clinical triage and diagnosis.\looseness=-1

Understanding engagement, its drivers and deterrents, is key to properly assisting essential workers and patients. We are interested in 
%want to detect and measure 
\emph{meaningful} engagement, %Contrary to the focus of engagement with other products or services, rather %than in its net increase, we are interested in understanding how to help %users achieve and maintain 
the optimal level of engagement that leads users to provide the best care to the communities they work with. Grasping and predicting engagement at the user level is a first step towards defining personalized interventions (such as adaptive learning journeys, suitable reminders, incentives for better practices, or drug recommendations) that can be delivered through the app just when they are needed. By boosting the performance of healthcare workers in greater need of support, these interventions can have a great impact on the community, in the form of improved care for patients.\looseness=-1
%By supporting specifically workers in resource poor settings, taking into account particularities such as offline app usage and high latency, these interventions could contribute to help alleviate global health inequities.

This work proposes different methods to study user engagement. Special attention is paid to understanding churn (user attrition) and its risk, understood as a marker of total disengagement. We pursue a multidimensional approach to the problem 
%that can provide insights on various use cases, 
and illustrate it by analyzing user activity from the \theapp{} \citetheapp, a digital training tool developed by \theorg{} \citetheorg\theotherorg{} that contains evidence-based obstetric and newborn guidelines for skilled birth attendants.\looseness=-1

The paper is organized as follows. In the remainder of this section, we describe related works and our contribution. In Section~\ref{sec:method}, we present our dataset and framework. Its application to the \theapp{} is discussed in Section~\ref{sec:results}, where various churn definitions are compared. Finally, in Section~\ref{sec:conclusion} we deliver our conclusions.\looseness=-1

\subsection{Related works}
\label{sec:related-works}

% TODO: ADD MORE REFS OR BLEND WITH INTRO. ANY NON-CHURN ENGAGEMENT REFS? ANY OTHER APPROACH TO CHURN RATHER THAN DAYS OF INACTIVITY? CHURN PREDICTION? REFS MHEALTH SOLUTIONS AND ANC?

Previous analyses of user engagement in digital health contexts can be found in \cite{Serrano2017}, where classification trees are employed to divide the user population %according to 
by
their different levels of engagement, and in~\cite{abdulsalam2022}, a survey on app abandonment. However, both studies focus on apps designed for the public, and thus not specific to healthcare workers. Other related studies propose mobile health solutions inspired on gaming research, or {\it health games}~\cite{Durga2015, Elnasr2015}. 

While there is limited work on general engagement, churn in non-contractual settings has been profusely studied for different products and services. Churn is usually defined in terms of a period of inactivity, which can be fixed (e.g. a calendar month with no logins) or rolling. For instance, in \cite{florian}, $35$ days without a gaming session, sport betting ticket or deposit are deemed evidence of churn in an online game, 
%\cite{cous} uses $4$ months of inactivity for online gambling, 
while \cite{lai} defines churn as no music downloads for a year. 
%and \cite{demetri} resorts to the quantiles of the days between transactions for a retail store.
In the context of video games, different methods to predict churn were explored in \cite{runge2015}, with other works using survival analysis~\cite{perianez2016, guitart2019} and time-series \cite{delrio2021} approaches. In the latter studies, the period of inactivity that determines churn is computed through the \emph{returning churners' method} (see Section~\ref{sec:return_churners_method}), which serves as a baseline for churn detection in this paper.

User behavior in the \theapp{} has been 
%previously 
studied in \cite{guitart2021}, which predicts the demand for specific in-app contents by certain groups of users, and in \cite{guitart2021b}, which explores the use of deep learning click-through-rate prediction models for content recommendation.\looseness=-1

\subsection{Our contribution}
To the best of our knowledge, this is the first time that probabilistic and engagement score methods are explored in the context of app engagement, or that time-to-churn is predicted through a survival ensemble method with time-varying covariates. We are not aware either of any previous work discussing various methods to measure user engagement and personalized ways of characterizing it.\looseness=-1

\section{Dataset and Methodology}
\label{sec:method}

%In this study, we propose a multifaceted approach to the study of user engagement. In the framework described, and depending on the use case, different methodologies can be applied to adequately selected activity indicators, focusing on one or some of the different angles under which the problem of engagement can be considered. Different methods to describe engagement lead, as will be soon described in more detail, to different possibilities of how to define churn risk \footnote{\emph{Risk} here is used in a generic way to describe the likelihood of a user having lost all interest in the app, rather than assuming its precise survival analysis definition.} (understood as total disengagement from the app), some of which will be more useful than other depending on the goal. Besides determining which activity indicators are the most suitable for the problem at hand, and which methods will be used to analyse these, the axis (or angles) along which to perform the analysis also need to be determined. The main possibilities to that effect considered throughout this paper as described below.

We propose a multifaceted approach to study user engagement, applying various methods to selected activity indicators 
%and to describe engagement, 
and different churn 
%\footnote{\emph{Risk} here is used in a generic way to describe the likelihood of a user having lost all interest in the app, rather than its precise survival analysis definition.}
%(understood as total disengagement from the app) 
risk definitions.
%, some of which will be more useful than others depending on the goal. 
%Besides this, 
The analysis can be performed from diverse angles, and we considered the following main dichotomies: 
%to that effect:

%exogenous: toda esta parte de los ángulos cada vez me queda menos claro si aporta valor, atl vez sea mejor asegurarnos de que esté bien explicado dentro de la metodología y resultados y quitar toda esta parte (o reducirla mucho al menos). osotros con ojos nuevos creo que juzgáis mejro si aporta lago o no
\emph{Frequency vs.\ intensity.} Engagement involves the frequency with which users perform certain actions (e.g. logging in), but also the intensity they devote to an activity (e.g. how long they are connected).\looseness=-1

%engage with a certain activity, for example, login, i.e., with how often and how recently the user has performed certain actions. Engagement, however, also has to do with the intensity with which the user is involved in a certain activity, for example, with how much time the user spends connected to the app.

\emph{Generic vs.\ specific.} Engagement can be defined in terms of generic metrics relevant to all apps, such as login frequency, number of sessions, 
time spent on the app 
or average number of clicks per session. 
%If we are interested in understanding distinct types of meaningful engagement depending on the purpose of app and the specific use case at hand, we might want to 
But we can also focus on app-specific measures of engagement, such as 
%those related to 
the time spent trying to solve quizzes in an e-learning app.\looseness=-1
%, or on how rapid the progression through the different content difficulty levels is.

\emph{Exogenous vs.\ endogenous.} To characterize the engagement of a given user, we can compare their present and past activities, using \emph{endogenous} (self-referential) indicators or scores. But we can also compare their activity to that of a group they belong to, and then their level of engagement is defined \emph{exogenously}.

\emph{Historical vs.\ snapshot.} In exogenous approaches, we can extend the comparison to all past behavior of the group (a historical perspective), or restrict it to the behavior of the group users at a certain moment (taking a snapshot of the engagement at that time). 
%Only in the case of exogenous approaches is this an option (users can only be compared to themselves using their past behavior).

%% Probably this is better left as methodological split and leave the rest as angles along which these can be employed
%\emph{Probabilistic vs indicator-based.} As will be described in more detail in the following subsections, different methods are proposed for measuring engagement. While some focus on characterising user behavior in probabilistic terms (e.g. how likely it is a user connects on a given a day, or is still active in the app 20 days from now), others build indicators that capture the level of engagement of the user.

\emph{Analytic vs.\ predictive.} Most of this work revolves around analytic measures that characterize user engagement from observed behavior. But the application of survival analysis to 
%generate predictions of 
predict
the probability that users will remain active in the future
%, which is further described in Section~\ref{sec:survival_method}, 
is also explored.\looseness=-1

%In the rest of this section, we present the dataset used to illustrate this work and describe the different methodologies we explored.

%%%%%%%%%%%%%%%%%%%%%%%%%%%%%%%%%%%%%%%%%%%%%%%%%%%%%%%%%%%%%%%%%%%%%%%%%%%%%%%%%%%%%%%%%
\subsection{Dataset}
\label{sec:data}

Our dataset comes from usage logs of the \theapp{} \citetheapp, a digital learning tool providing skilled birth attendants with up-to-date evidence-based clinical guidelines,
%, including the core components of Basic Emergency Obstetric and Newborn Care. 
%The effectiveness of the \theapp{} in helping midwives 
%and other skilled birth attendants 
%to acquire and maintain critical skills 
whose effectiveness
has been evaluated in clinical trials~\cite{Lund2016, Oladeji2022}. The content of the App is divided into clinical modules, which are each comprised of educational videos, easily referenceable action cards, drug list, and practical procedures, as well as a gamified learning platform containing a series of tests of increasing difficulty to assess the users’ acquired knowledge and skills.
%and knowledge acquired.%(maybe we can remove this or shorten?)

The analyzed dataset comprises $58195$ users from India ($95\%$) and Ethiopia ($5\%$) and shows their in-app activity between January 2018 and August 1, 2021. The user activity logs were processed into daily user metrics characterizing daily user behavior and engagement. 

%For example, Figure \ref{fig:daily_ts} gives an example of one of this user metrics. In particular, it depicts the e-learning connection time, or time in seconds spent daily using e-learning features of the app (videos, action cards and testing), for two selected users of the \theapp{}, to which we will refer as user 1 (in blue) as user 2 (in red). The behavior of these two users, and the implications in assessing their level of engagement, will be revisited in the discussion of results in section \ref{sec:results}. % (and in the supplementary material of the appendix \ref{appendix}).

%\begin{figure}
%  \centering
%    \includegraphics[width=\linewidth]{user_conn_time.png}
%  \caption{Time series of daily user metric of connection time spent in e-learning (in seconds) for two users.}
%  \label{fig:daily_ts}
%\end{figure}

The specific daily metrics we considered include lifetime (number of days between first login and a certain moment), daily connection time, 
%(daily time spent using the app), 
action count (number of clicks), e-learning action count (number of clicks on videos, action cards and testing features), progression (number of tests successfully passed), video view count (number of videos watched), video watch time,
%(time in seconds spent watching videos), 
loyalty index (fraction of days with login), and days since last login. 
%(days of difference between day considered and last login date, has minimum value 1). 

The analysis 
%of this dataset 
in Section~\ref{sec:results} assumes all this user history is available, and explores what it tells us about the engagement of individual users and of the user population as a whole on the last observed date (August 1, 2021).\looseness=-1 

%TODO: MENTION OTHER PLOTS IN SUPPLEMENTARY IF ANY, OR REMOVE CURRENT ONES IF WE DISREGARD APPENDIX

\subsection{Returning churners and missed metrics} 
\label{sec:return_churners_method}

The approaches to define churn risk presented in this work will be compared to a \emph{baseline method} previously used in the context of video games~\cite{perianez2016, guitart2019}, which considers that a user has churned after $k$ consecutive days without login. The goal of this method is to find a \emph{churn definition}, i.e. a reasonable way of setting the value of $k$. 

%While this is not a method to study engagement generally speaking, we describe here the method used to determine when users churn used in \cite{perianez2016, guitart2019} (in the context of free-to-play videogames), as it will be used as a baseline to which to compare the additional methods to define churn risk that will be presented in this work. As in most of the previous churn related research described in Section~\ref{sec:related-works}, a user is considered churned 

The idea is to choose a value as small as possible, while satisfying constraints on the number of users we {\it wrongly} classify as churners. %and the activity coming from these users upon their return. 
This is tantamount to setting thresholds on the fraction of false or \emph{returning churners} 
%(i.e., fraction of all the users designated as churned that will log back again in the future) 
(users identified as churners that nonetheless will log in again in the future). 
%\footnote{These are referred to as \emph{false churners} in earlier papers on this methodology~\cite{perianez2016, guitart2019}}. 
%If we are interested in also enforcing that these returning churners will not contribute significantly to any activity that we consider a particularly important engagement marker after they have been deemed as churned by the method, additional thresholds can be applied on relevant \emph{missed metrics} (fraction of a particular metric coming from returned churners).
We can also enforce that these returning churners will not contribute significantly to any considered activity by setting additional thresholds on relevant \emph{missed metrics} (the fraction of a particular metric coming from returned churners).\looseness=-1

There is always a tradeoff between accuracy and efficacy when selecting a specific churn definition: the longer we wait 
%for a user to remain inactive before we label them 
before we label inactive users
as churned, the more certain we can be that they have actually quit. %However, this information will be of little use, as the connection to the user will have been lost long ago. 
However, quickly detecting churn is more useful, as attempts to reengage the user are more likely to succeed.

%CUT: Aquí se puede quitar bastante, lo más importante es el final (It is probabilistic in the sense that...). Sí se puede dejar algo de lo anterior fenómeno, si no con eso vale
In terms of the angles discussed 
%Note that when considering the angles introduced 
at the beginning of this section, such a churn definition is analytic (as it relies exclusively on observed behavior), historical and exogenous (as it considers the past behavior of large groups of users). It always uses a generic measure of frequency (through the threshold on returned churners), that can be combined with additional constraints on generic or specific measures of intensity (through thresholds on missed metrics). It is {\it probabilistic} in the sense that it constrains the probability of {\it wrongly} classifying a user as churned, but it does not involve probabilistic statements about individual user behavior, as the empirical cumulative function (ECDF) or survival methods described below.
%in Sections~\ref{sec:probabilistic_method} and \ref{sec:survival_method}.

%The resulting 
The specific churn definitions 
%when applying this method to our dataset %described in Section~\ref{sec:data} are 
%presented and 
for our dataset
are
discussed in Section~\ref{sec:results-return-churners}. The missed metrics considered are connection time, action count, and progression. 
%%% JCHECK: Changed "e-learning connection time" as per Babaniyi's comments
%%% BABANIYI: Checked - it's correct. Please change "e-learning action count" to "action count"
%%% JAVIER: Done
And we set thresholds of 30\% and 10\% on returning churners and missed metrics, respectively.\looseness=-1

%%%%%%%%%%%%%%%%%%%%%%%%%%%%%%%%%%%%%%%%%%%%%%%%%%%%%%%%%%%%%%%%%%%%%%%%%%%%%%%%%%%%%%%%%%%%%%
\subsection{Empirical cumulative distribution function}
\label{sec:probabilistic_method}

Probabilistic approaches use the available distributions of observed behavior to find the likelihood of a certain activity level for a user, as well as their resulting engagement. 
%%%JCHECK: Check if the previous sentence is correct or should be modified
%as described by that particular daily activity. 
%%% BABANIYI : Checked - it's correct.
The ECDF method 
%described below 
is an analytic probabilistic approach, 
%(see Section~\ref{sec:survival_method} for a predictive probabilistic approach). 
where we can consider the distribution for (i) the same user in the past (ECDF endogenous, ECDF-endo); (ii) the group they belong to (ECDF exogenous, ECDF-exo), or (iii) that group on a particular day (ECDF snapshot exogenous, ECDF-snp). 
%%%JCHECK: Check "daily user behavior distributions" (and the whole sentence)
Daily user behavior 
%daily metric 
distributions do not fall clearly into a parametric family of distributions and tend to be left-skewed, and thus we will focus on non-parametric methods, which make no underlying assumptions about the probability distribution. 

The \emph{cumulative distribution function} (CDF) is defined as $F(x) = P(X \leq x) = \sum_{t\leq x} f(t)$ for a discrete random variable $X$, and as $F(x) = \int_{-\infty}^{x} f(t) dt$, where $F(x)$ is a non-decreasing continuous function, for a continuous random variable 
%$X$ 
(which here corresponds  
%$X$ would be here an 
to a specific activity metric).
%, which can be generic or specific, and intensity or frequency related. 
Given $N$ ordered data points $y_{1}, y_{2}, \dots, y_{N}$, the \emph{empirical cumulative distribution function} (ECDF) takes the form $E_{N} = {n(i)}/{N}$, where $n(i)$ is the number of points less than $y_{i}$, and the $y_{i}$ are arranged in ascending order. This is a step function that increases by $\frac{1}{N}$ at each ordered data point and converges to the CDF when enough data is collected. 
%%% JCHECK: This definition of the ECDF seems to convey it is a number divided by another number, which is likely not true.
%%%% BABANIYI: For the ECDF definition, it is divided by the total number of points. I think we should also change the notation from P to N to be consistent.
%%%% BABANIYI: Given $n$ ordered points $y_{1}, y_{2}, \dots, y_{n}$, $E_{n}=  {n(i)}/{n}$ and the step function $\frac{1}{n}$ 
%%%% JAVIER: Changed, but I kept capital N for the total number of points, as think it is better to use different symbols for n(i) and N.
The ECDF indicator is defined as the value of the ECDF at time
%for the day considered 
$t$, i.e., the probability with which that level of activity or less was to be expected:
%CUT: tal vez dejar las equaciones dentro del texto? Hay pocas pero no es que sean complejas
\begin{equation}
\label{eq:ecdf_indicator}
    e_{t}^{i} = F^{i}(z_{t}^{i}\,|\,z_{1}^{i}, \ldots, z_{t-1}^{i} ),
\end{equation}
where $F^{i}$ is the empirical cumulative distribution function for the user $i$, and $z_{t}^{i}$ is the user metric that reveals engagement. Note that we are considering the user's behavior as compared to themselves, and hence taking an endogenous approach.

%CUT: Aquí podemos dejar explícitamente sólo el primer ejemplo y quitar los de la última parte
For example, let us consider a frequency-related feature typically used in this analysis: the days between a (generic or specific) activity for a single user. 
%(that we will compare to themselves). 
We intuitively know that anomalously large values are pointing to disengagement and potential churn, so we can treat the situation as an anomaly detection problem. The ECDF indicator allows us to make claims like \textit{$9$ times out of $10$, a user will perform this activity again within $z$ days} (for an ECDF indicator value of 0.9 at~$z$). If the period of inactivity extends beyond $z$, we can thus deem it an unusually low engagement for that user.
%, as there was only a $10\%$ chance of the happening.  
%of $1$ in $10$, 
For intensity metrics, we can build an intensity ECDF indicator that will allow us to make similar claims, such as 
%around the intensity of the activity by interpreting the percentile of interest accordingly. For example, 
\textit{$9$ times out of $10$, a user will correctly reply to less than $x$ test questions daily} (where an ECDF indicator value of 0.9 describes a very engaged testing behavior). %or \textit{there is a $10\%$ chance a user will spend at least $x$ minutes on the App}. 
%(for an ECDF indicator value of 0.1).

The previous statements refer to the likelihood that a user shows a certain level of activity as compared to their past behavior %of that same user. 
By considering the ECDF distribution for all users, the statements would become about how usual or unusual user behavior is as compared to the conduct of their group---their country, in our case---, either in general or on a certain day (in the snapshot approach).\looseness=-1

%To derive estimates of the nature described above, we calculate the time between activities and take a non-parametric approach by using the ECDF to approximate the quantiles of the days between activity distribution. Once we have the ECDF, we approximate any percentile threshold (e.g. $90^{th}$ in this case) we deem appropriate. Similar reasoning can be applied to intensity features such as connection time, leading to claims such as \textit{$9$ times out of $10$, a user will correctly reply to at least $x$ test questions daily} or something like \textit{there's a $90\%$ chance a user will spend at least $x$ minutes on the app}. Note these statements refer to the likelihood of a certain user level of activity as compared to themselves. By considering the ECDF for all users, the statements would become about how usual or unsual user behavior is as compared to the group (in general or on a certain day if we only consider the observations for that day.)

The results described in Section~\ref{sec:results-probabilistic} consider exclusively a frequency measure, namely days between logins, looking at it from endogenous, historical exogenous, and snapshot exogenous perspectives. These combined give us an indication 
of the likelihood of the user's time since last login.
%of how likely last login recency is
%as compared to their previous login history, to that of all users from their %country, and to that of users from their country on August 1, 2021.

To study churn detection using these ECDF approaches, we can set a threshold on the likelihood of a certain engagement trait, and consider users inactive (i.e. at least temporarily churned) if the level of activity is deemed unusually 
%unlikely 
low. Here (see Section~\ref{sec:results-churn-detection}) we consider any observed behavior with a probability equal or greater than 0.1 (the $10$ or $90$ percentiles described in the examples above) \emph{not} suspicious of inactivity. If an unusually low activity has 
%only 
%%% JCHECK: I removed "only" because it did not make sense to me
%%% BABANIYI: Checked - it's correct
been observed in the past with a likelihood of less than 0.1, it is considered to signal high churn risk.

The ECDF approach offers a very flexible way to understand engagement along different angles of interest, yielding a quite intuitive interpretation of both measures of engagement and churn definitions. Its main limitation is that, even if it can be used to provide a collection of measures characterizing the engagement of a user on a given day (in general and for specific activities, as compared to themselves or to their group, either on that day or in the past), these are not easily reconciled into a single unified measure. % or churn definition. 

%%%%%%%%%%%%%%%%%%%%%%%%%%%%%%%%%%%%%%%%%%%%%%%%%%%%%%%%%%%%%%%%%%%%%%%%%%%%%%%%%%%%%%%%%%%%%%%%%%%%
\subsection{Engagement scores}
\label{sec:engagement_method}

Another analytic approach to characterize user engagement is to build \emph{engagement scores} (which typically range between 0 and 1).
%and that characterize the user's level of engagement. 
Different engagement measures (intensity and/or frequency related, generic and/or specific) can be combined into a single indicator by taking their harmonic mean:
%CUT: tal vez dejar las equaciones dentro del texto? Hay pocas pero no es que sean complejas
%\begin{equation}
%\label{eq:engagement_indicator}
%    s_{t}^{i} =
%    \dfrac{n}{\sum\limits_{j=1}^{n} 
%    \dfrac{1}{\hat{z}_{jt}^{i}} }
%\end{equation}

%\begin{equation}
%    s_{t}^{i} = n \Big/
%    \sum\limits_{j=1}^{n} 
%    \frac{1}{\hat{z}_{jt}^{i}} 
%\end{equation}
\begin{equation}
\label{eq:engagement_indicator}
    s_{t}^{i} =
    \dfrac{n}{\sum\limits_{j=1}^{n} 
    {\left(\hat{z}_{jt}^{i}\right)^{-1}} }
\end{equation}
where $n$ is the number of metrics combined into the indicator, and $\smash{\hat{z}_{jt}^{i}}$ are the scaled metric component values for user $i$ at time $t$. 

To keep the score bounded between 0 and 1, we apply min-max scaling to all features used to build the indicator. The scaling is performed as compared to the past values of the same user if we want that component of the score to reflect endogenous engagement, or as compared to the past group of users (introducing thus exogenous historical components), or to the behavior of a group of users at time $t$ 
% on that day
%%% JCHECK: I replaced "on that day" with "at time $t$", for consistency. 
%%% BABANIYI: Checked - it's correct
(exogenous snapshot components). Note that whenever one of the component metrics is zero, the engagement score goes to zero, and thus it will always vanish on days without activity if we are using intensity measures. 
%is not well defined. 
%This should be taken into account when considering this method for churn detection, by combining it with a frequency indicator (for example, by considering the average of a frequency and an intensity score). 

%Imagine we want to build a generic personalised engagement comprising connection time and action time (both intensity measures) score. If by applying min max scaling to connection time and action time for each user, the scaled values for the last observed daily activity of a given user are $0.35$ and $0.7$, then the score for that user on that day will be $e_{t}^{i} = 2/(1/0.35 + 1/0.7) = 0.4667$. This would be signaling an average level of engagement as compared to the previous activity of the user. In particular, a lower than usual time spent using the app is compensated by a larger than usual number of actions performed during that time. 

%CUT: No consideramos este método para detección de churn risk explícitamente en los resultados, así que esra referencia se puede reducir a la mínima expresión 
%As with all the other approaches to quantify engagement, if we want to use the engagement score 
%approach 
%to determine
%high 
%churn risk, we need to set a threshold to determine what level of unusually low engagement will be deemed a signal of inactivity or churn. 
The greatest strength of this approach is its ability to combine components of different types,
%\footnote{While this possibility is not considered in this work, various components can be given different weights if we want to gauge their relative contributions to the final score.} 
which allows us to discriminate non-meaningful engagement even in days with activity. It uses a single unified quantification of engagement, thus leading to a single engagement measure that can be, however, based on different metrics. %The method's 
The
main weakness lies in the difficulty of having an intuitive understanding of what the score represents and how it relates to churn.\looseness=-1

The engagement score used in Section~\ref{sec:results-score} is built from the following user metrics: weekly loyalty index, video view count, video watch time, action count, progression, and e-learning connection time. 
%While it will not be discussed explicitly in 
%the results in 
%Section~\ref{sec:results-churn-detection}, 
One could also set thresholds on the score for churn risk detection (e.g. considering all users with scores between 0 and 0.1 are in high risk of churn, even if they have activity on that day).

\subsection{Survival analysis}
\label{sec:survival_method}

Survival analysis can also be used to characterize user engagement, and is the only predictive approach considered in this work. Survival 
%analysis 
models seek to predict the time to an event of interest, and output the probability that such event has not occurred yet as a function of time. Hence, if we set the models to predict time-to-churn, we obtain the probability that each user is still actively using the app as time goes by.
%, i.e., their expected disengagement with time. 
Survival analysis is particularly well-suited to deal with censored data, meaning that, while for many users we do not know the time-to-churn because they are still active, these models can learn from the fact that these users have not churned yet (as opposed to standard regression 
%modeling 
approaches).

This methodology is frequency-based in that login frequency over time is what determines the user lifetime, understood as the days between their first and last logins. Nonetheless, it incorporates information on both intensity and frequency through the use of both types of metrics for feature engineering, and both generic and specific quantities can be considered. This approach is historical and exogenous by definition (models learn about the expected behavior from the past observed behavior of a group of users).

%Besides consider the time measured in days since first login, we can use other cumulative quantities, such as for example, accumulated time spent using the app. 
%While this methodology does not explicitly consider measures of engagement frequency and intensity and the focus is on churn prediction, by combining predictions in days since first login with modeling in an accumulated intensity variables (such as connection time), some idea on how frequently and intensely app use will be in the future is conveyed. While this approach is well suited to work with generic engagement indicators, it can also work with more specific measures of {\it time}, such as in levels progressed or exams attempted. This approach is historical and exogenous by definition (models learn about the expected behavior from the past observed behavior of a group of users). 

An additional decision that needs to be made when using this method is what definition of churn we will use, in order to identify whether the event of interest occurs or not. In this paper, a churn definition of 31 days is used for both Ethiopian and Indian users. This is just a convenient choice, and it can be modified depending on the use case. It represents roughly a month, and is also, as we will see, what the ECDF exogenous approach described in Section~\ref{sec:probabilistic_method} points to (see Section~\ref{sec:results-churn-detection}). This means users will be labeled as churned after 31 consecutive days without logging in. This method can thus never be used to define churn, as it relies on an already existing churn definition. However, it can be combined with any other of the methods precisely by using the churn definition they provide. %In the discussion of this work in section \ref{sec:results-survival} a churn definition of 31 (as determined by the ECDF-exo method applied to days between logins, see table \ref{tab:churn-def-compare}) is used. 
In Section~\ref{sec:results-survival} we will discuss how to use the survival curves to characterize user
%behabior
engagement, 
and the accuracy of the models in predicting churn.

\begin{figure}
  \centering
  \includegraphics[width=\linewidth]{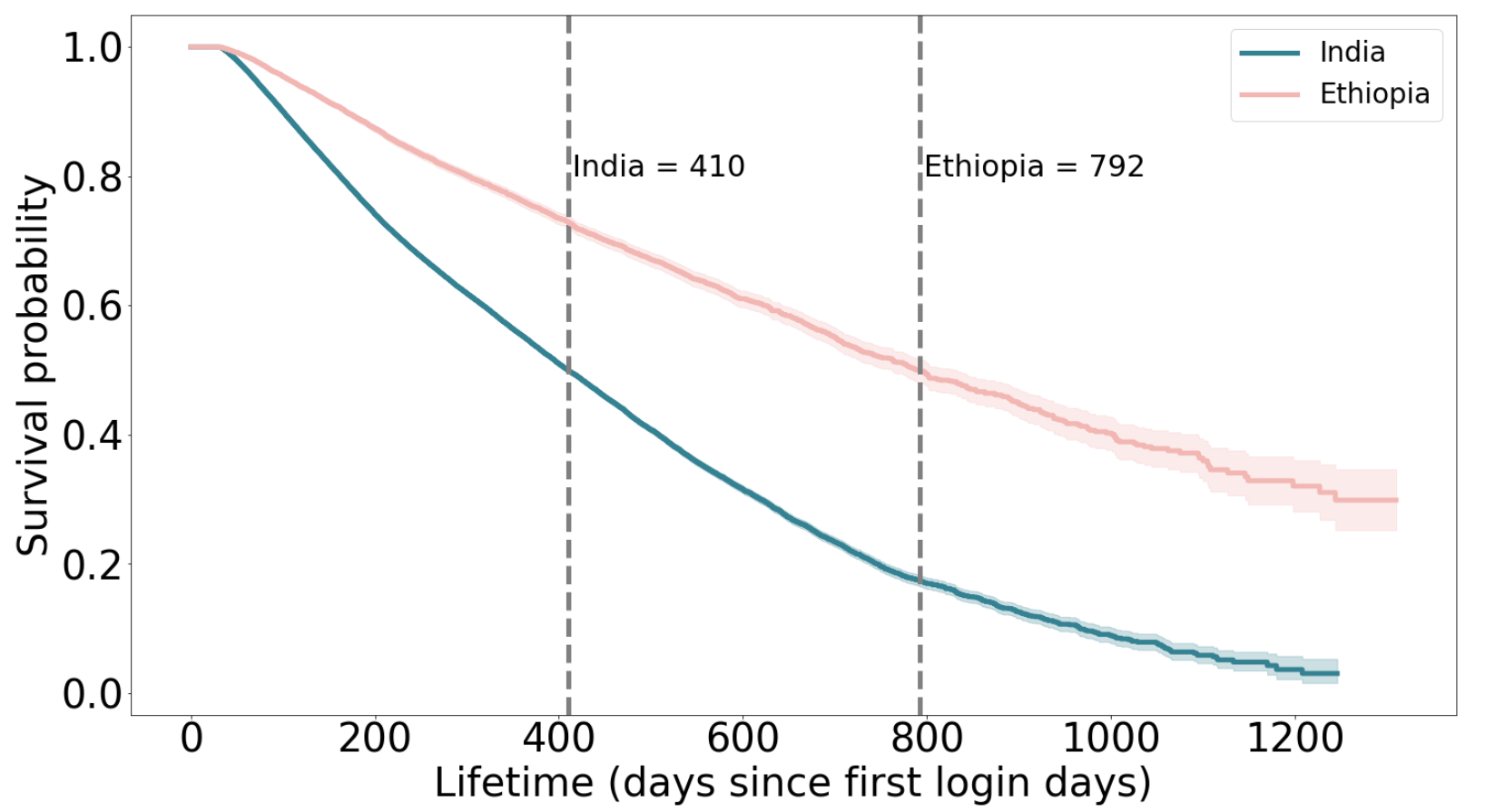}
  \caption{Kaplan--Meier estimates
 %in days since first login 
 for India and Ethiopia. Shaded areas show 95\% confidence intervals.}
    \label{fig:km}
\end{figure}

Figure \ref{fig:km} shows the Kaplan--Meier estimates 
%in days since first login 
for the whole user population in India and Ethiopia. It depicts the fraction of users that were still active (according to our definition) in the studied dataset as a function of days since first login. And it shows median lifetime is nearly twice as large for Ethiopian users (792 days vs.\ 410 days in India). This is consistent with the fact that the App was introduced earlier in Ethiopia. The Indian population comprises more users that employ the App for a short time when they learn about it or during their studies, while a more significant fraction of the Ethiopian users already have a lasting relationship with the App.\looseness=-1

We focus on two survival analysis models: conditional inference survival ensembles (CSF) and their extension to time-varying covariates, given by left-truncated and right-censored (LTRC) models. We evaluate the model performance at all available times using the integrated Brier score (IBS), 
%evaluation metric, proposed by \cite{Graf}. The \textit{IBS} 
which 
%provides an overall estimate of the model performance at all available times and 
is a number between $0$ and $1$, with $0$ being the best possible value. It makes use of the Brier score (BS), which estimates the mean squared distance error between the observed event outcome and the predicted survival probability.

%\begin{equation}
%    IBS (t_{max}) = \dfrac{1}{t_{max}} \int_{0}^{t_{max}} BS(t)dt
%\end{equation}

%where $t_{max}$ is the maximal time for estimating the prediction error curves and $BS(t) = \dfrac{1}{N} \sum_{i=1}^{N}(1_{T_i > t} - \hat{S}(t, \overrightarrow{x_{i}}))^{2}$.

To develop good predictive models (IBS close to 0), we created additional time-series features, such as week of the year, connection time, action counts, or session counts in the past 3, 7 and 15 days. We then fit the CSF model on $b=25$ bootstrap rounds (this is done by taking multiple samples with replacement from a single random sample),
%as suggested by \cite{efron}. The choice of $b=25$ rounds 
a number chosen to minimize computation time. Subsequently, we calculate the average IBS as ${\rm IBS_{boot,avg}} = \frac{1}{b}\sum_{j=1}^{b=25} {\rm IBS}_{{\rm boot},j}$. For each bootstrap round, the model was trained using $75\%$ of the users' data and validated on the remaining $25\%$. In the LTRC case, we proceed similarly: for the time-varying covariates, we use relevant engagement metrics (as discussed above) and time windows that capture their variation over a day, a week, or a month. Lastly, we use the top $30$ features with the highest feature importance scores across all the bootstrap rounds to fit the final CSF and LTRC models analyzed in this study.

%TODO: ADD KM PLOTS STRATIFIED BY COUNTRY, MODEL SPECIFICATIONS, DETAILS ON FEATURE ENGINEERING

%CUT: Llegado el caso podemos suprimir las subsecciones y dejarlo explicado con el resto del texto, aunque yo diría que como último recurso
%\subsubsection{Conditional Inference Survival Ensembles (CISE)}
\subsubsection{Conditional Survival Forests (CSF)}
%TODO: ADD KM FORMULA

This is an ensemble method that recursively partitions the feature space to maximize the difference between the survival curves of users belonging to different nodes~\cite{Hothorn2006, CISE}. The split is performed in terms of Kaplan--Meier estimates. CSF models avoid the split variable selection bias (favoring of splitting variables with many possible split points) present in random survival forests~\cite{Ishwaran2008} by using linear rank statistics as the splitting criterion at each node.

\subsubsection{Left-Truncated and Right-Censored (LTRC) Forests}\label{sec:LTRC}

The LTRC conditional inference forest (LTRC-CIF) model~\cite{Fu2016, Yao2020EnsembleMF} is similar to the static feature CSF, but can handle left-truncated data and %an extension of the static feature CSF model to handle left-truncated data (besides the right-censoring already described) and
%, very particularly, 
time-varying covariates. The LTRC relative risk forest (LTRC-RRF)~\cite{Fu2016, Yao2020EnsembleMF} is another model apt to deal with right-censored data, an adaptation of the relative risk forests discussed in~\cite{LeBlanc1992} to time-varying covariates. Conceptually, in both LTRC models, the time-varying covariates are used to generate different pseudo-user observations out of observations of the behavior of a single user. While the LTRC-CIF model performs the split using Kaplan--Meier estimates, the LTRC-RRF resorts to the Nelson--Aalen estimator.

%TODO: ADD SOFTWARE USED SUBSECTION WITH REFS TO LIBRARIES AND LINK TO GITHUB REPO, OR LIB REFERENCES IN TEXT AT LEAST TO PYSURVIVAL AND LTRC

%\subsection{Software used}

%%%%%%%%%%%%%%%%%%%%%%%%%%%%%%%%%%%%%%%%%%%%%%%%%%%%%%%%%%%%%%%%%%%%%%%%%%%%%%%%%%%%%%%%%
%%%%%%%%%%%%%%%%%%%%%%%%%%%%%%%%%%%%%%%%%%%%%%%%%%%%%%%%%%%%%%%%%%%%%%%%%%%%%%%%%%%%%%%%%
\section{Results and discussion}
\label{sec:results}

In this section, we apply our framework to study user engagement among the Indian and Ethiopian users of the \theapp{}.
%We try to illustrate how the methods can be used and combined to give a a multidimensional characterization of the engagement of specific selected users at a given moment in time, but also how this user analysis for the whole population can be considered to characterize engagement in the latter. We consider as the day of interest the final day with data in dataset, August 1, 2021, so as to replicate a production environment where we want to assess user and population engagement as described by the latest (and previous) information available.
We try to illustrate how the discussed methods can provide a multidimensional description of the engagement of specific users at a given time, but also how they can characterize the engagement of the whole population.
%%% JCHECK: Check the previous sentence
We take the last day with data, August 1, 2021, as the day of interest, to replicate a production environment where one wants to assess user and population engagement in terms of the latest information available.

As the methods described can be applied to different metrics or metric combinations, and through different angles (as discussed at the beginning of Section~\ref{sec:method}), the possibilities are literally endless and need to be carefully chosen for the specific use case at hand. 

To demonstrate our approach, we have focused on:
\begin{enumerate}
    \item The ECDF indicator through different angles---endogenous (ECDF-endo), historical exogenous (ECDF-exo), and snapshot exogenous (ECDF-snp)---for a generic frequency metric (days between logins). In exogenous approaches, the group will always consist of all the users from the same country.
%%% JCHECK: Added as per Babaniyi's comments
    \item The engagement score (ES) combining frequency and intensity metrics, both generic and specific, with only endogenous components.
    \item Survival curves in days since first login, as predicted by a collection of frequency and intensity, generic and specific features (and comparing three different models).\looseness=-1
\end{enumerate}
%The discussion is illustrated by 
%considering and 
%comparing the results of applying 
The results of applying these measures of engagement to two selected users of the \theapp{} (to whom we will refer as users 1 and 2) on August 1, 2021, are summarized in Table~\ref{tab:user-results}. In the next sections, we will present additional figures and information to illustrate the kind of discussion allowed by this approach.
%, where for the selected two users the different measures of engagement are compared for . Figures and information will be presented in the next subsections to illustrate the kind of discussion this approach allows for. %Additional plots and statistics for a few additional users to support the analysis presented here are available in the appendix as supplementary material for the interested reader.

%TODO: MENTION ADDITIONAL SUPLEMENTARY MATERIAL IF WE INCLUDE ANY

%TODO: COMPLETE TABLE!  

\begin{table}
\caption{Various engagement measures for two selected users on August 1, 
%2020.
%%% JCHECK: I changed the year to 2021
2021. 
ECDF-endo-CD gives the equivalent churn definition for the users as per their ECDF-endo, and SurP is the survival probability predicted by the CSF model.}
    \label{tab:user-results}
    \centering
    \small\setlength{\tabcolsep}{3pt}
    \begin{tabular}{@{}llllllll@{}}
    \toprule
    User & ECDF-endo & ECDF-exo & ECDF-snp & ES & ECDF-endo-CD & SurP \\
    \midrule
%, 175     
    1 & 0.98 & 0.99 &1 & 0.029 & 29 days & 0.82 \\
    2 & 0.76 & 0.55 & 0.29 & 0.331 & 6 days & 0.69\\
    \bottomrule
    \end{tabular}
\end{table}

%%%%%%%%%%%%%%%%%%%%%%%%%%%%%%%%%%%%%%%%%%%%%%%%%%%%%%%%%%%%%%%%%%%%%%%%%%%%%%%%%%%%%%%%%
\subsection{Returning churners and missed metrics}
\label{sec:results-return-churners}

\begin{figure}
  \centering
  \includegraphics[width=\linewidth]{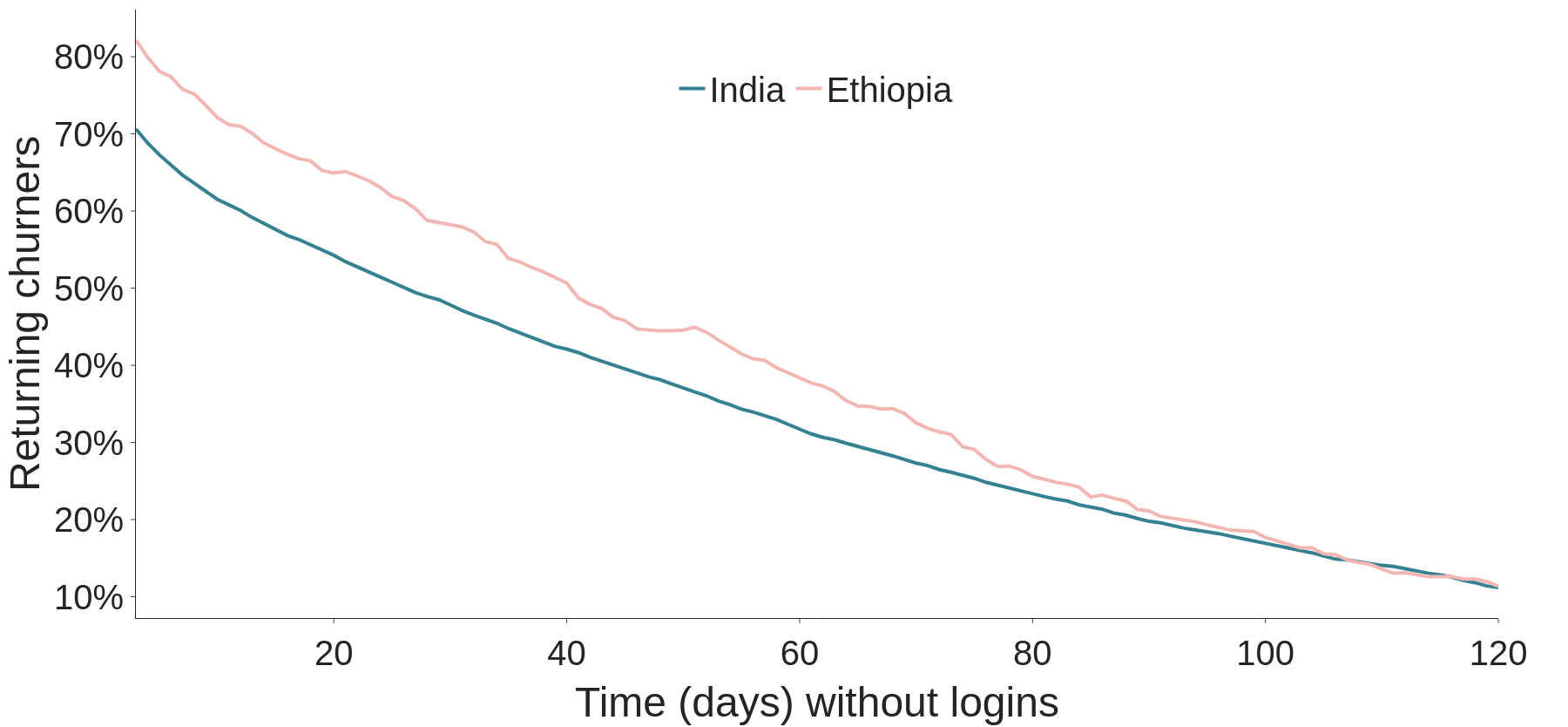}\\
    \includegraphics[width=\linewidth]{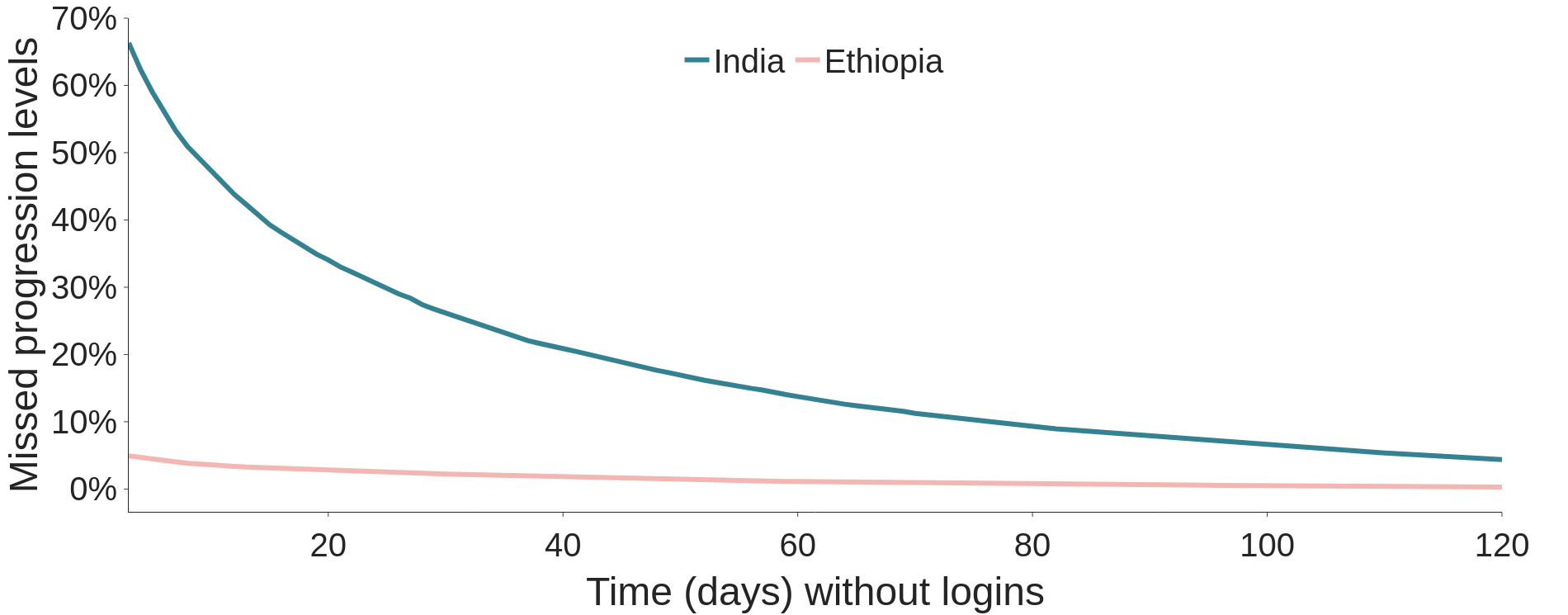}
  \caption{Fraction of returning churners ({\itshape top}) and missed progression ({\itshape bottom}) 
  %vs.\ churn definition 
  for India ({\itshape green}) and Ethiopia ({\itshape pink}), when we define churn as $k$ consecutive days without logins.}
    \label{fig:return_churners}
\end{figure}

%We present the results of the return churners and missed metrics methods explained in Section~\ref{sec:return_churners_method} for 
To find a suitable churn definition, in
Figure~\ref{fig:return_churners} we plot the percentage of returning churners and of missed progression (tests successfully passed by returning churners after they came back) as a function of
$k$ (the number of consecutive days without logins after which a user is deemed a churner).   
%values of the churn definition. 
For relatively small $k$, the probability of wrongly classifying a user as a churner is larger for Ethiopia, so one needs to consider longer periods of inactivity to spot disengagement among Ethiopian users. 
%time than Indian users to signal disengagement. 
However, the difference among countries is small, diminishes with $k$, and vanishes after about three months. In contrast, the behavior of the missed progression (and all other e-learning %specific 
missed metrics considered) is markedly different for both countries. Returning churners do not make any meaningful contribution to the overall App progression in Ethiopia, so it would be justified to adopt a shorter churn definition even if there are many returning churners. India's curve, however, is consistent with users significantly engaging in learning activities (successfully taking tests) in bursts separated by relatively long pauses.

If we choose the churn definition that will keep the fraction of returning churners below 30\% and of relevant missed metrics (connection time, action count and progression) 
%%% JCHECK: Changed "e-learning connection time" as per Babaniyi's comments
%%% JAVIER: Changed "e-learning action count" to "action count", as per a previous comment.
below 10\%, we find that $k$ should be 74 days for Ethiopia and 64 days for India. (Different thresholds on different metrics can be chosen depending on the specific use case.) Note that these churn definitions are hardly actionable: if we need to wait over two months to decide a user has quit, it will be almost impossible to reengage them, as they will have probably lost all their interest long ago. Therefore, in Section~\ref{sec:results-churn-detection}, we will explore the possibility of combining this method with the ECDF approach in order to detect churn risk much earlier. The baseline churn definition, however, remains
%%% JCHECK: Check if "The baseline churn definition" at the beginning of this sentence is correct.
the most robust way of unequivocally spotting users who are not going to come back (the situation we seek to minimize).

%Table~\ref{tab:return_metrics} has been included in the supplementary material included in the appendix (section \ref{app:retrun-churners}) and presents suitable churn definition for different thresholds of returning churners and missed metrics.

%CUT: es interesante porque muestra un comportamiento distinto para los distintios países, pero podemos suprimer la gráfica de missed progresión (y o bien dejar la expliacaión en el texto, o si es necesario, eliminar también de ahí la comparación para ganar unas líneas)

%%%%%%%%%%%%%%%%%%%%%%%%%%%%%%%%%%%%%%%%%%%%%%%%%%%%%%%%%%%%%%%%%%%%%%%%%%%%%%%%%%%%%%%%%
\subsection{ECDF indicator}
\label{sec:results-probabilistic}

\begin{figure}
  \centering
  \includegraphics[width=\linewidth]{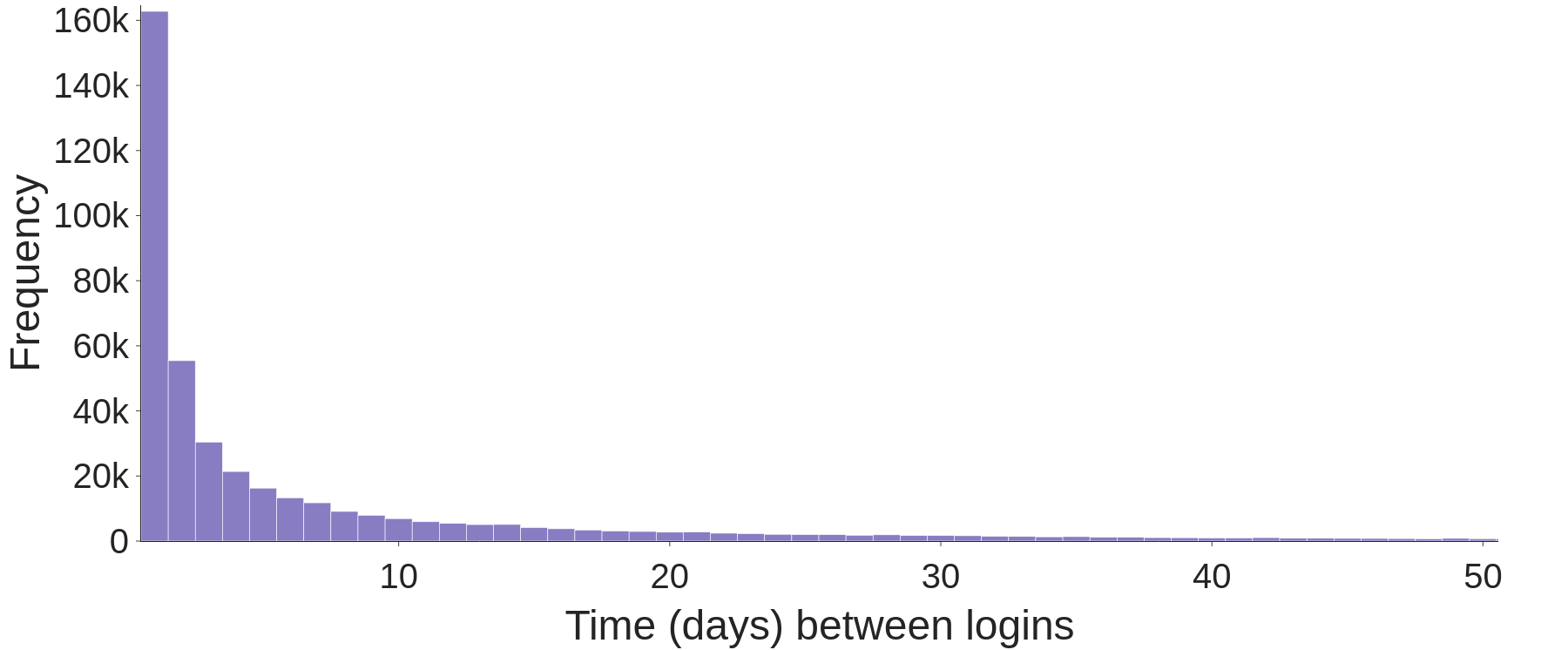}
  \caption{Histogram of days between logins for all users in India and Ethiopia, considering all historical data.}
    \label{fig:time_login_dist}
\end{figure}

We applied the ECDF method to the frequency metric of days between logins, whose histogram is depicted in Figure~\ref{fig:time_login_dist}. We enforce a cutoff at 200 days to limit the bias and noise introduced by the arbitrarily large values that churners would keep contributing. Note that the most typical pattern is a few days between logins, and most instances are below two weeks. However, there is also a very long tail, implying that some users will log in again after many months of inactivity. This distribution would be used to compute the user ECDF under the exogenous historical (ECDF-exo) approach.

%CUT: Prefiero que no pero si hay que sacrificar otra figura yo diría que esta

%TODO: ADD TS PLOTS AND DESCRIPTION

\begin{figure}
  \centering
  \includegraphics[width=\linewidth]{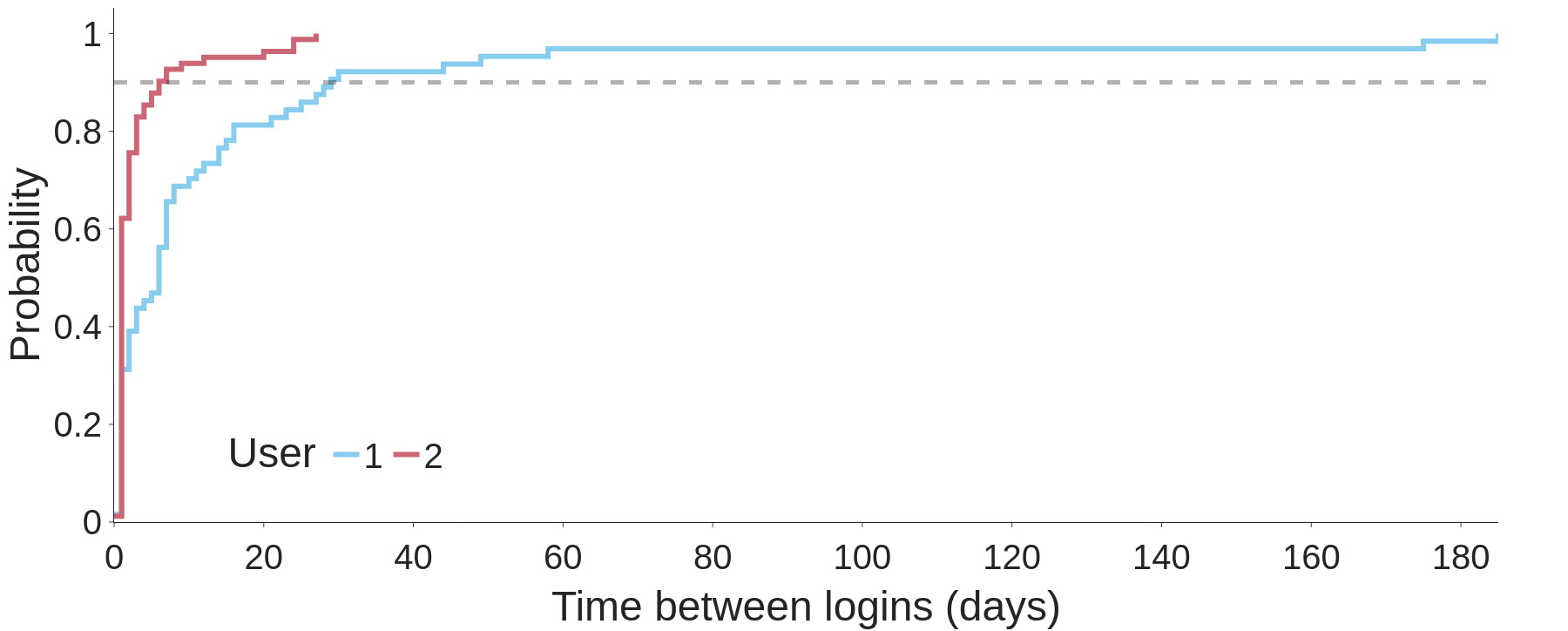}
  \caption{Endogenous ECDF distribution of days between logins for users 1 ({\itshape blue}) and 2 ({\itshape red}). The 0.9 probability horizontal line marks the {\itshape churn alert} threshold.}
    \label{fig:prob-ecdf}
\end{figure}

%Figure \ref{fig:time_login_users} shows the same distribution for user 1 (blue) and user 2 (red). These will be the distributions used for the ECDF endogenous historical (ECDF-endo) approach to build their ECDFs, which are depicted in figure \ref{fig:prob-ecdf}. %compares the endogenous ECDF for the two users%, while figure \ref{fig:prob-ts} depicts the daily values of this endogenous ECDF measure for the same users (additional users and plots corresponding to the application along exogenous angles are included in the supplementary material). 
Figure~\ref{fig:prob-ecdf} shows the ECDF-endo cumulative distribution of days between logins for our two selected users. The graph has a much longer tail for user 1, implying a larger uncertainty as regards the number of days typically expected between logins. User 2 logs in more frequently and consistently, with almost no instances of more than 30 days between sessions, while user 1 displays a more erratic behavior, with longer periods between logins (up to 6 months). % that can reach up to 
%longer average periods of time between logins of up to 2 months, and occurrences for much longer periods (close to 6 months). 
%6 months.\looseness=-1
%\begin{figure}
%  \centering
%  \includegraphics[width=\linewidth]{user_time_between.png}
%  \caption{Histogram of days between logins for user 1 (blue) and user 2 (red).}
%    \label{fig:time_login_users}
%\end{figure}

%TODO: INCLUDE PLOT COMPARING INDIA AND ETHIPIA DISTRIBUTIONS

%Figure \ref{fig:prob-dist} compares how this endogenous ECDF measure of engagement is distributes across user populations in India and Ethiopia for the [DATE] (additional plots corresponding to the application along exogenous angles are included in the supplementary material in section \ref{app:prob} of the appendix). Note how...

%TODO: COMMENT ON TABLE VALUES 

For those two users, the August 1, 2021, values of the three ECDF engagement indicators discussed in Section~\ref{sec:probabilistic_method} are presented in Table~\ref{tab:user-results}. 
%These are the  ECDF endogenous historical value (ECDF-endo, comparing user engagement to their own activity history), the ECDF exogenous historical one (ECDF-exo, comparing user engagement to all the previous activity of the group of users) and the ECDF exogenous snapshot one (ECDF-snp, comparing user engagement to the rest of the users of the group for that same day). 
%Note the systematically 
The very high values of all ECDF indicators for user 1 result from a long period of inactivity before the day considered. 
%(August 1, 2021, when the user did log in). Indeed, 
Indeed, the last login prior to August 1 happened 175 days before for user 1 (and 2 days before for user 2). The indicator values for user 2 show this user is particularly engaged as compared to the rest of users on that day (ECDF-snp), pretty disengaged as compared to their own past activity (ECDF-endo), and moderately engaged when taking into account the whole group's history (ECDF-exo).  

Figure~\ref{fig:prob-ecdf} also includes the 0.9 threshold we set to infer a high risk of inactivity or churn. Note that the use of this particular metric (days between logins) from an endogenous perspective is analogous to finding an equivalent churn definition for each user,
%(dynamically evolving with time),
similar to that obtained through the returning churners' method. The values of $k$ corresponding to that equivalent definition (which are also included in Table~\ref{tab:user-results}) are 29 days for user 1 and 6 days for user 2. This means that, in the observed history of each user, nine out of ten times they have logged in again after less than those days. For user~2, a week without logins already points to a lower-than-expected engagement, while in the case of user 1 we need to wait for almost a month to conclude the same. In contrast, the exogenous approach results in a fixed churn definition for the whole user population, as will be discussed in Section~\ref{sec:results-churn-detection}.

\begin{figure}
  \centering
  \includegraphics[width=\linewidth]{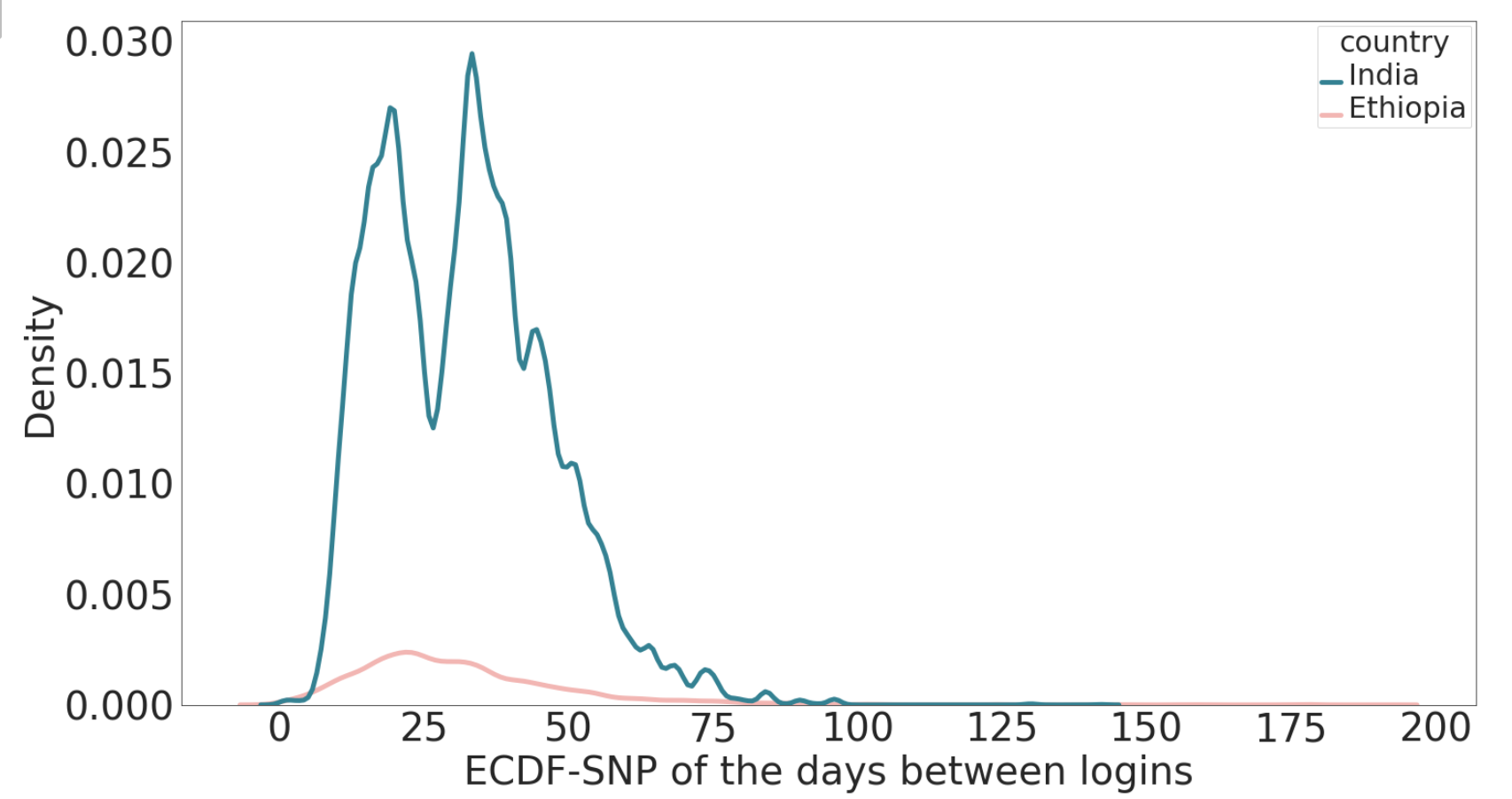}
  \caption{Distribution (kernel density estimation) of ECDF-snp 
%indicator 
values for August 1, 2021, across India and Ethiopia.  }
    \label{fig:snp-dist}
\end{figure}

Figure \ref{fig:snp-dist} shows the values of the ECDF-snp indicator on August~1, 2021, across India and Ethiopia. The marked bimodal behavior for India indicates that%, among the users who logged in on that day, 
there are two large groups whose last login took place roughly 20 and 40 days earlier. 

%TODO: REFERENCE ANY ADDITIONAL PLOTS IN SUPPLEMENTARY MATERIAL

%%%%%%%%%%%%%%%%%%%%%%%%%%%%%%%%%%%%%%%%%%%%%%%%%%%%%%%%%%%%%%%%%%%%%%%%%%%%%%%%%%%%%%%%%
\subsection{Engagement score}
\label{sec:results-score}

\begin{figure}
  \centering
  \includegraphics[width=\linewidth]{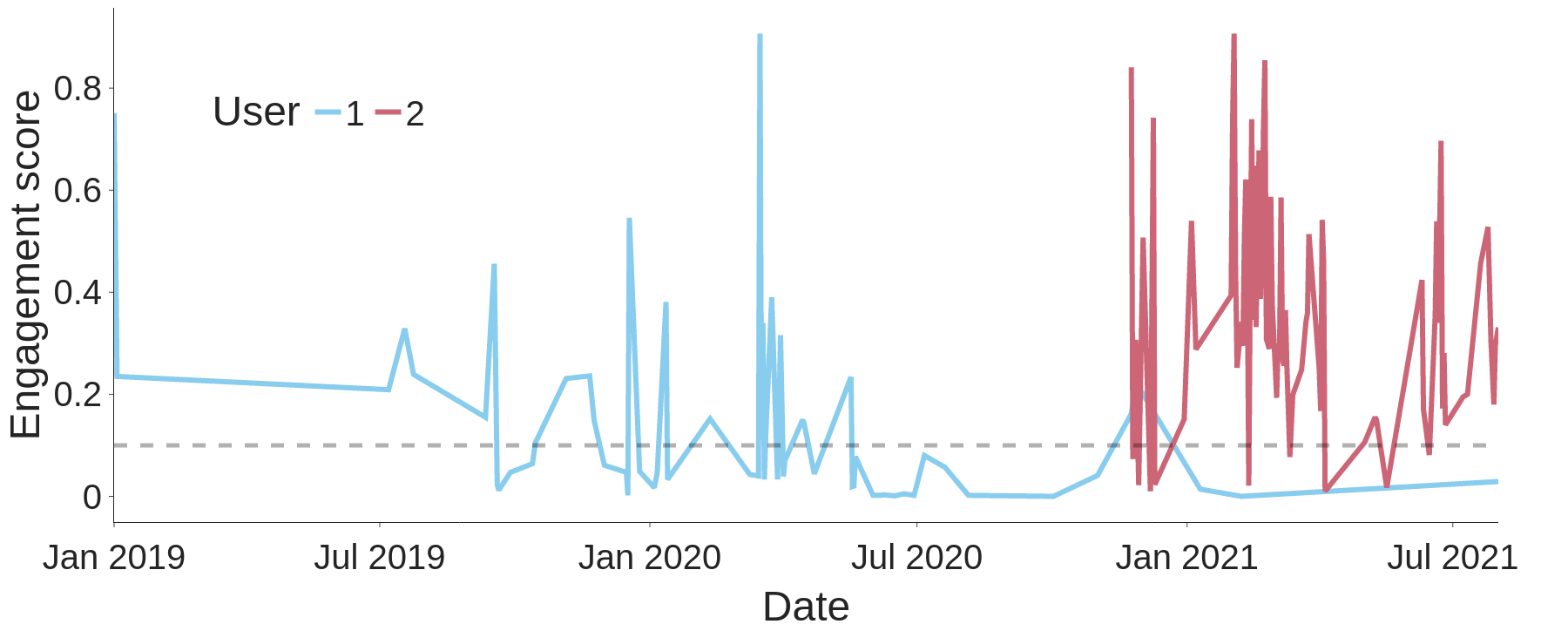}
  \caption{Time series of endogenous daily engagement scores for user 1 ({\itshape blue}) and user 2 ({\itshape red}). The 0.1 probability horizontal line marks the {\itshape churn alert} threshold.}
    \label{fig:user-scores}
\end{figure}

As described in Section~\ref{sec:engagement_method}, we built an endogenous engagement score relying on weekly loyalty index, video view count, video watch time, action count, progression and e-learning connection time. Figure \ref{fig:user-scores} compares the time series of the daily values of this engagement score for the two selected users. Larger values of the engagement score correspond to a higher level of engagement, as the metrics considered \emph{increase} with engagement (contrary to the days between logins discussed in the previous section). The engagement score for August 1, 2021, is included in Table~\ref{tab:user-results} and shows that---as compared to their own past behavior---user 2 is moderately engaged, while user 1 is at risk of churn, even if they registered some activity on that day. Therefore, this method also captures that user 2 is more engaged than user 1, just as the ECDF indicators.\looseness=-1
%(even as compared to themselves).

%CUT: Esto podría ir fuera
%The daily time series scores for the two users show how these capture the bursts in intensity of both users, while the inclusion of the weekly loyalty index will contribute to raise or low the value for average intensity days depending on the frequency of login of a given a week. This particular endogenous engagement score we are considering here will boost/downgrade the engagement associated to any particular activity depending on how recent and frequent the latest activity has been.

%CUT: Si se ha explicado esto en la metodología igual se puede quitar esto de aquí (o redecirlo mucho)
Figure \ref{fig:user-scores} also depicts the 0.1 threshold that marks an unusually low engagement, indicative of a high risk of churn. While we will not explicitly consider the engagement score for churn detection, it could be used to that end in the same way as the ECDF indicators. However, when discussing churn, we prefer to focus on the frequency ECDF approach as, by construction, it will yield values more easily comparable to our baseline churn definition (since both refer to consecutive days without logins), besides covering the endogenous, exogenous and snapshot angles. 

\begin{figure}
  \centering
  \includegraphics[width=\linewidth]{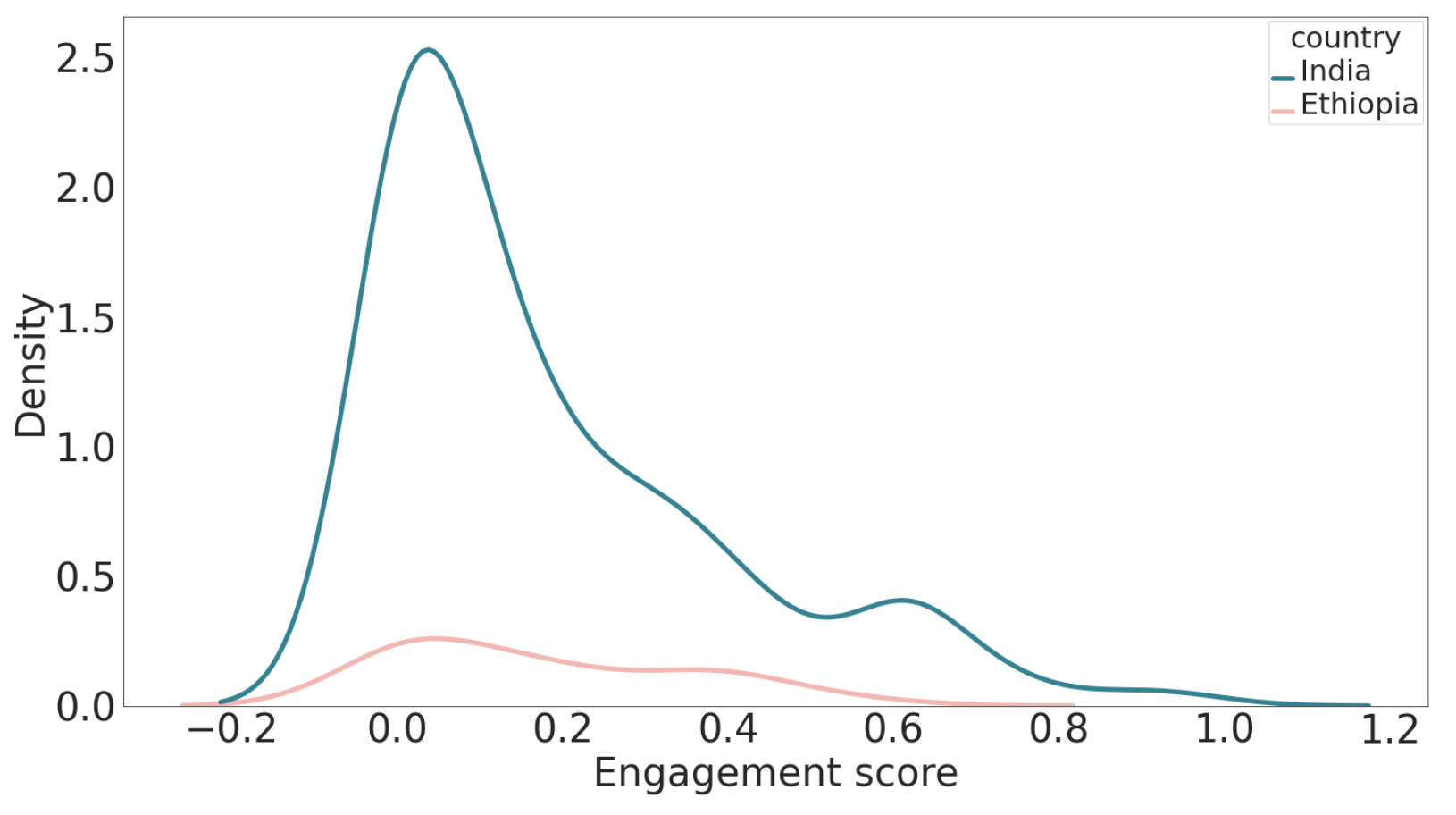}
  \caption{Distribution (kernel density estimation) of engagement scores for August 1, 2021, across India and Ethiopia.}
    \label{fig:scores-dist}
\end{figure}

Figure \ref{fig:scores-dist} compares the distributions of engagement scores across the user populations of India and Ethiopia for August 1, 2021. While both graphs are similar, India's is more skewed to the left, indicating a more disengaged population (as compared to their own past behavior) on that day. 
%as compared to the Ethiopian users. 
This is consistent with a longer history of use for most Ethiopian users, which makes the current ones relatively loyal. Both India and Ethiopia distributions are bimodal, with a large peak associated to relatively low engagement, and another smaller concentration of 
%probability from 
more engaged users. Note that this 
%group around the more engaged mode, 
second group is more engaged (higher engagement score) in the Indian case.\looseness=-1

%%%%%%%%%%%%%%%%%%%%%%%%%%%%%%%%%%%%%%%%%%%%%%%%%%%%%%%%%%%%%%%%%%%%%%%%%%%%%%%%%%%%%%%%%
\subsection{Churn detection}
\label{sec:results-churn-detection}

We have discussed other ways of detecting churn besides our baseline method (whereby a user has churned after a certain number of consecutive days without logins, found by imposing thresholds on the percentage of returning churners and missed metrics). These alternative definitions are based on the ECDF endogenous approach (where we set a 0.9 probability threshold on the distribution of days between user logins) and both exogenous ones (where we compare the {\it typical} days between logins for a user and for the group,%the whole population, 
either historically or on the day of interest, also using a 0.9 threshold). Note that any churn definition relies on determining a meaningful level of activity, so that users below a certain threshold are considered to be inactive.

%: churn definition (that can also be measured in days since last login) as given by the 0.9 threshold on ECDF endogenous approach, and both exogenous ones (comparing the {\it typical} days between logins of users to that of the whole population historically and on the day of interest). 
%Another alternative that will not be considered here explicitly is applying the 0.1 threshold to the endogenous engagement score proposed (made of both frequency and intensity components). 

\begin{figure*}
  \centering
  \includegraphics[width=\linewidth]{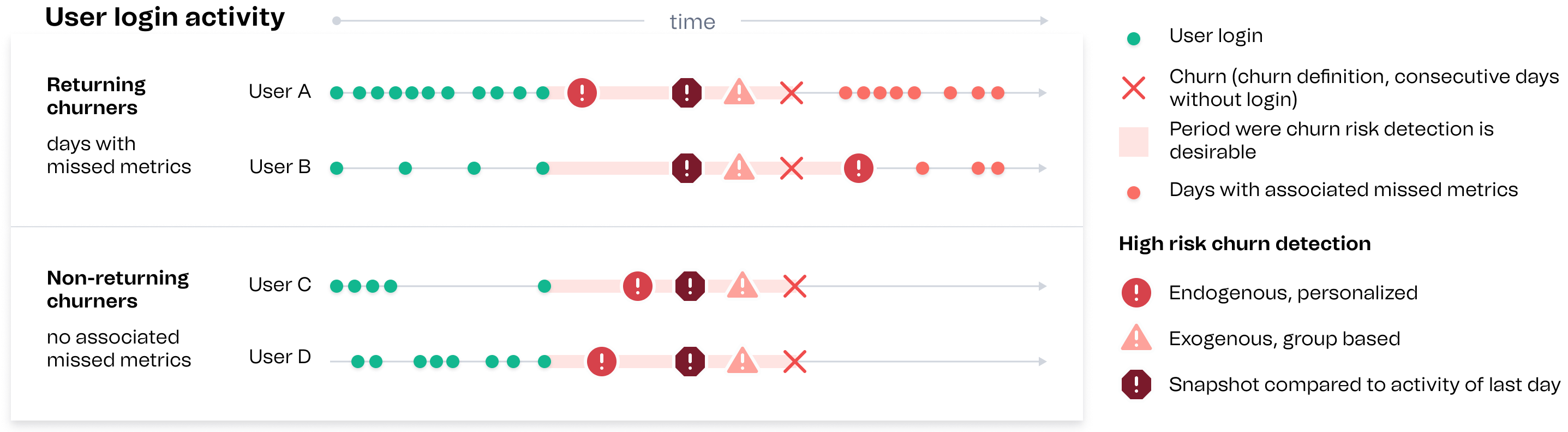}
  \caption{Schematic representation of the different churn detection methods employed in this study.}
    \label{fig:churn-schema}
\end{figure*}

The use of the ECDF equivalent churn definitions to complement the returning churners' baseline one is schematically represented in Figure~\ref{fig:churn-schema}.
%, and its implications are then discussed for the dataset under study. 
The diagram shows user activity (days with logins are represented as green or red dots), and the day when churn is detected using the baseline method is marked with a red cross (according to that method, all users churn at the same time). 
%This is always after 
%%%%%churn definition 
%a number of
%consecutive days without login, with the churn definition set to comply with the required thresholds in returning churners and missed metrics. 
In the diagram, users 3 and 4 are examples of returning churners, and activity corresponding to the days after they were marked as churned 
%they logged in 
(red dots), will contribute to missed metrics. The ECDF-indicator-based churn definitions will typically detect high risk of churn (through atypically disengaged behavior) earlier. The precise moment is marked with exclamation marks, surrounded by a circle, hexagon, and triangle for the endogenous, snapshot, and exogenous indicators, respectively. The two latter methods still detect churn at the same time for all users. The former, however, yields a personalized churn definition, which will identify churn much earlier for very engaged users---but perhaps later than the baseline churn definition, as in the case of user 3, for users with a low engagement pattern. % as compared to their peers.\looseness=-1

Table \ref{tab:churn-confusion-matrix} presents the confusion matrix as of August 1, 2021. That is, it shows the number of users 
%that both 
classified as churned/not churned by the RCMM approach, and in high/low risk of churn according to the ECDF indicators (endogenous, exogenous and snapshot) threshold method. 
%, as non-churned and in no churn risk by both, and more interestingly, the extra potential churners detected by either method. 
Note that the ECDF-exo approach finds 20 additional users in high risk of churn (for whom the number of days since their last login is larger than the number of days observed 90\% of the time, for all users and all times) who are not identified by the RCMM method. This is also the case for 16 users when considering the self-referential engagement (ECDF-endo) indicator. Interestingly, there are 12 users who are deemed churners by our baseline method (which roughly means they have not logged in within the previous two months) while appearing significantly engaged as compared to their own past activity, as per the ECDF-endo indicator. 
%still capturing as significantly engaged as compared to themselves.

\begin{table}
  \caption{Number of users labelled as \textit{churned} and \textit{not churned} by the ECDF-endo and ECDF-exo methods as compared to the returning churners and missed metrics (RCMM) baseline for August 1, 2021, across India and Ethiopia.}
  \label{tab:churn-confusion-matrix}
  \centering\small
  \begin{tabular}{lllll}
    \toprule
     & \multicolumn{2}{c}{ECDF-endo}  & \multicolumn{2}{c}{ECDF-exo}                  \\
    \cmidrule(lr){2-3}\cmidrule(lr){4-5}
    RCMM       & Not churned     & Churned & Not churned & Churned\\
    \midrule
    Not churned   & 223            & 16    & 220   & 20   \\
    Churned       & 12            &  6   & 0   & 18  \\
    \bottomrule

  \end{tabular}
\end{table}

As we applied the ECDF method to a metric that is frequency-based and equivalent to detecting churn through a number of consecutive days without logins (as in our baseline approach), we can compare the resulting equivalent churn definitions. Note that, since exogenous methods characterize churners as {\it inactive as compared to the group}, the equivalent churn definition will be the same for all users. In contrast, in the ECDF endogenous approach, each user has their own churn definition to mark unusually low inactivity. Table~\ref{tab:churn-def-compare} shows the churn definition values arising from the different methods considered (in the ECDF-endo case, the average is presented) as of August 1, 2021. Note that, for more engaged populations, we normally need shorter periods of time to detect churn (if the login frequency is high, relatively short periods of no activity already signal disengagement).

The equivalent churn definitions that we found are consistent with the picture already described. Note that the baseline method yields the stricter definition, as a longer period of inactivity is needed to conclude that a user is not going to come back. On the other extreme are the average values for the ECDF-endo method---although we know some users may have churn definitions longer than the baseline ones, as for the 12 users discussed above. As for the remaining methods, the snapshot approach yields greater engagement (lower churn definition) than the exogenous one among Ethiopian users, while the contrary is true for India. 
This means that, in India, the period of time needed to detect churn is significantly longer now than in the past (disengaging population), while Ethiopian users appear slightly more engaged nowadays. 

When comparing both countries, Ethiopian users are more likely than Indian ones to use the App again after long periods of inactivity (as indicated by the longer RCMM definition). The average engagement of users as compared to their own past activity (ECDF-endo) and that of their compatriots (ECDF-exo) is very similar in both countries. However, Indian users are significantly more engaged than their Ethiopian counterparts when considering only August 1, 2021, as reference (ECDF-snp).

%Figure \ref{fig:churn-personalized-dist} depicts the distribution of endogenous churn definitions in India and Ethiopia. The averages are also included in table \ref{tab:churn-def-compare}, together with the (single for both exogenous methods) churn definition yielded by the comparison to all past behavior of all users of the same country, and to that of the day of interest. Note...

%TODO: DISTIRBUTIONS OF PERSONLISED CHURNDEF IN ETHIOPIA/INDIA

\begin{table}
    \caption{Churn definitions (number of consecutive days without logins needed to determine churn or high churn risk) for India and Ethiopia, calculated using the returning churners and missed metrics (RCMM), ECDF-exo, ECDF-snp, and ECDF-endo methods. (For the latter, average values are given.)}
    \label{tab:churn-def-compare}
    \centering\small
    \begin{tabular}{@{}lllll@{}}
    \toprule
    Country & RCMM & ECDF-exo & ECDF-snp & ECDF-endo (Avg)  \\
    \midrule
    Ethiopia & 74 & 29 & 20  & 20 \\
    India & 64 & 31 & 52 & 19 \\
    \bottomrule
    \end{tabular}
\end{table}

\subsection{Survival Analysis}
\label{sec:results-survival}

Predictions of the survival probability---the probability of not having churned yet, i.e., of having logged in within the previous 31 days (where the number of days is
%%% JCHECK: Check if "the number of days" is correct.
%%% ANA: checked
determined by applying the ECDF exogenous approach to the days between logins)---as a function of days since first login (lifetime) are calculated using conditional inference ensembles, as well as the two LTRC forest models with time-varying covariates described in Section~\ref{sec:LTRC}. % (plots for additional users are available in the supplementary material in section \ref{app:survival} of the appendix), and the values of the survival probability for [DATE] included for all selected users in table \ref{results-users}, as well as the current and remaining predicted lifetimes and connection times. After $b=25$ bootstrap rounds, $IBS_{bootavg} = 0.133 \pm 0.0075$. Since there's negligible difference in the $IBS_{boot,j}$ as evident in the standard deviation $(0.0075)$, we evaluated the feature importance at each bootstrap round and use the best $30$ features across all rounds to create the final CISE and LTRC models. For the final models, we trained the model using the data from $75\%$ of the users and validated its performance on the remaining $25\%$.

\begin{figure}
  \centering
  \includegraphics[width=\linewidth]{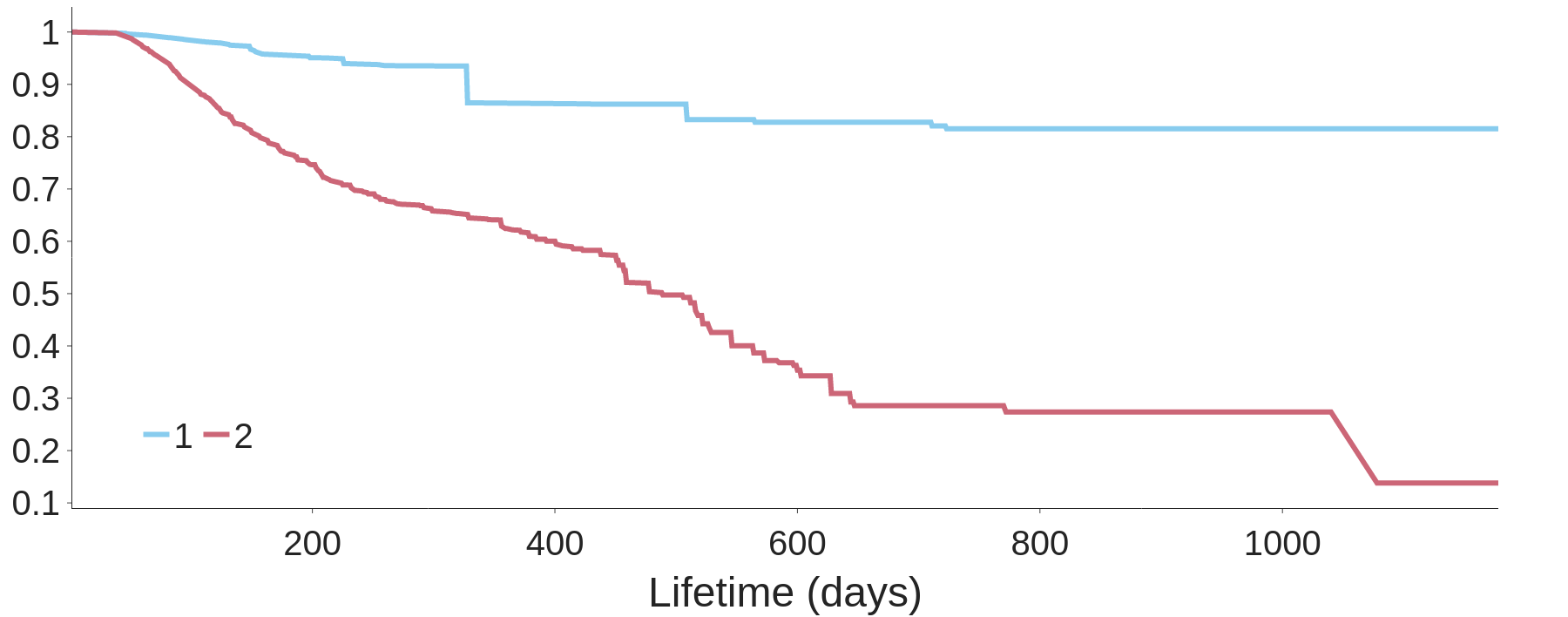}
  \caption{Probability of not having churned for users 1 ({\itshape blue}) and 2 ({\itshape red}) as a function of days since first login, as predicted by CSF lifetime model. }
    \label{fig:survival-curves}
\end{figure}

An example of the resulting survival curves for the selected users is depicted in Figure~\ref{fig:survival-curves}. These represent the predicted probability of each of the users not having churned (i.e., of having logged in at least once in the previous 31 days). The actual predicted lifetime would be given by the median value. While user 2 is shown to be more engaged than user 1 by all our analytical measures,
%%% JCHECK: I added "analytical" and swapped users 1 and 2 (originally it said user 1 was more engaged than 2)
%%% BABANIYI: Checked - It's correct. However, the probability of being active for user2 is 0.69 not 0.26 while for user1 it's 0.82, please correct
%%% JAVIER: Done
this user is also predicted to disengage relatively quickly and quit around 250 days after first login. However erratic and less consistent was user 1's behavior 
%they might have displayed 
in the past, their survival probability never falls below 0.5, meaning that (with the available data) the model does not predict this user will ever churn. This is reasonable if we consider that this user has been active for most of the available data period and has always returned, even after long periods of inactivity. Note that, as we are looking at lifetime predictions, we are considering frequency of connection rather than intensity. The survival probability for users 1 and 2 on August 1, 2021, is also included in Table~\ref{tab:user-results}. As we have just discussed, the not-too-engaged user 1 still has a survival probability significantly larger than 0.5, while the moderately engaged user 2 only has a 0.69 probability of still being active 
at that point (even if they \emph{are} actually active).
%at this point, even if they are active on that day and therefore definitely not churned according to our definition. 
 
%%% JCHECK: Please, check the last sentence, as I simplified it a bit.

Table \ref{tab:ibs} collects the IBS scores of the different models. Note that, as expected, the accuracy of both LTRC forests is better than that of the CSF model, due to their ability to deal with time-varying covariates. Among the two, the LTRC-RRF performs best.%approach is the best option in terms of accuracy. %(additionally, Brier Score plots and observed vs predicted plots are included with the supplementary material in section \ref{app:survival} of the appendix).

\begin{table}
  \caption{Integrated Brier Score for time-to-churn predictions.}
  %, where time is given in days since first login.}
  \label{tab:ibs}
  \centering\small
  \begin{tabular}{lllllll}
    \toprule               
    Model      &  Ethiopia      & India    \\
    \midrule
    CSF        & $0.130$ & $0.090$     \\
    LTRC CIF    & $0.101$ & $0.057$     \\
    LTRC RRF    & $0.092$ & $0.049$     \\
    \bottomrule
  \end{tabular}
\end{table}

%\begin{table}
%  \caption{Integrated Brier Score (\textit{IBS}) for the time to churn predictions, with time in days since first login (lifetime, left) or in accumulated connection time (right)}
%  \label{tab:ibs}
%  \centering
%  \begin{tabular}{lllllll}
%    \toprule
%     \multicolumn{3}{r}{Lifetime}  & \multicolumn{3}{r}{Connection time}                   \\
%    \cmidrule(r){2-6}
%    Model      &  Ethiopia      & India   &  Ethiopia      & India \\
%    \midrule
%    CISE        & $0.130$ & $0.090$ & $0.07 $      & $0.080 $     \\
%    LTRC CIF    & $0.101$ & $0.057$ & $0.167 $     & $0.156 $      \\
%    LTRC RRF    & $0.092$ & $0.049$ & $0.174 $     & $0.169 $      \\
%    \bottomrule
%  \end{tabular}
%\end{table}

% The model performance as measured by the \textit{IBS} when the survival time is lifetime is presented in Table \ref{tab:lifetime} while for accumulated connection time is presented in Table \ref{tab:accum_connection_time}.

\section{Summary and Conclusions}
\label{sec:conclusion}

We have proposed three methods to study user engagement in mHealth applications intended for frontline healthcare workers, and tested them with data from the \theapp{}, a digital training tool for skilled birth attendants. These methods can be applied to a variety of engagement-related metrics (dealing with frequency or intensity, generic across apps or use case-specific) in order to understand user behavior as compared to their past behavior, that of the whole group, or the population engagement on a particular day. This provides a very flexible and versatile multidimensional framework to explore engagement and churn, with the downside of yielding limitless possibilities and combinations. Additionally, the two analytic approaches can be used to detect churn by setting appropriate thresholds and comparing them to the 
%churn detection arising from the 
baseline method considered (based on constraints on the fraction of returning churners and missed metrics). We also added a predictive dimension to this study by considering survival analysis models to foresee time-to-churn in days since first login. 

This results in a holistic, well-rounded approach to study engagement and churn, which can be customized to suit the specific needs of different mobile apps and use cases. Both the ECDF and score methods are flexible and versatile, allowing us to incorporate different engagement indicators and angles to the analysis. While ECDF approaches yield an intuitively understandable measure of user engagement, 
%(as compared to themselves, or their peers, now or throughout the app's history), 
it is not trivial to use them to produce a single, unified measure of engagement (and hence a single magnitude in terms of which to define churn). The engagement score approach has the enormous advantage of providing such a single measure of engagement, 
%to use across the population and to define churn, 
but is not as easy to interpret.
%in turn does not share the easy to understand intuition under the measures yielded by the ECDF methodologies.

Survival analysis yields the expected user disengagement profile. It incorporates different engagement indicators through the use of features, providing additional insights that help to characterize a user's engagement, given by the survival probability at a certain time as compared to their previous history and expected future. Furthermore, it serves to predict when each user will churn, an additional quantity that can be used to describe engagement.

The methods presented in this work may serve to enhance engagement among health workers using online learning and capacity building tools. This, in turn, could translate into improved care for their patients and have a significant impact on global health.  

\begin{acks}
The authors thank Javier Grande for his careful review of the manuscript and Marisa Ansari for her design of the churn detection schema. This work was supported, in whole or in part, by the Bill \& Melinda Gates Foundation INV-022480. Under the grant conditions of the Foundation, a Creative Commons Attribution 4.0 Generic License has already been assigned to the Author Accepted Manuscript version that might arise from this submission.
\end{acks}

\section*{Data and code availability}
All data used in this analysis comes from the \theapp{} logs and belong to \theorg{}. For inquiries regarding its use, please contact them at mail@maternity.dk. Code used is available at https://github.com/benshi-ai/paper\_2022\_kdd\_engagement.%, together with a collection of synthetic daily user metrics. (instead of those corresponding to the \theapp{}).

%\begin{acks}
%The authors thank Javier Grande for his careful review of the manuscript. This work was supported, in whole or in part, by the Bill & Melinda Gates Foundation INV-022480. Under the grant conditions of the Foundation, a Creative Commons Attribution 4.0 Generic License has already been assigned to the Author Accepted Manuscript version that might arise from this submission.
%\end{acks}

\bibliographystyle{ACM-Reference-Format}
\bibliography{main}

\end{document}